\documentclass[runningheads]{llncs}
\usepackage{graphicx}

\usepackage{tikz}
\usepackage{comment}
\usepackage{amsmath,amssymb} 
\usepackage{color}
\usepackage{algorithm}
\usepackage{algorithmic}
\usepackage{wrapfig}
\usepackage{float}
\usepackage{subcaption}
\usepackage{graphicx}
\usepackage{multirow}
\usepackage{booktabs}
\usepackage{adjustbox}
\usepackage{enumitem}
\usepackage{arydshln}

\usepackage[accsupp]{axessibility}  

\begin{document}
\pagestyle{headings}
\mainmatter
\def\ECCVSubNumber{4}  

\title{Object Detection in Aerial Images\\ with Uncertainty-Aware Graph Network} 

\titlerunning{Object Detection in Aerial Images with Uncertainty-Aware Graph Network}
%

\authorrunning{J. Kim et al.}
%

\author{Jongha Kim\inst{1} \and
Jinheon Baek\inst{2} \and
Sung Ju Hwang\inst{2,3}}

\institute{Korea University\and KAIST\and AITRICS\\
\email{jonghakim@korea.ac.kr, \{jinheon.baek,sjhwang82\}@kaist.ac.kr}}


\maketitle

\begin{abstract}
In this work, we propose a novel uncertainty-aware object detection framework with a structured-graph, where nodes and edges are denoted by objects and their spatial-semantic similarities, respectively. Specifically, we aim to consider relationships among objects for effectively contextualizing them. To achieve this, we first detect objects and then measure their semantic and spatial distances to construct an object graph, which is then represented by a graph neural network (GNN) for refining visual CNN features for objects. However, refining CNN features and detection results of every object are inefficient and may not be necessary, as that include correct predictions with low uncertainties. Therefore, we propose to handle uncertain objects by not only transferring the representation from certain objects (sources) to uncertain objects (targets) over the directed graph, but also improving CNN features only on objects regarded as uncertain with their representational outputs from the GNN. Furthermore, we calculate a training loss by giving larger weights on uncertain objects, to concentrate on improving uncertain object predictions while maintaining high performances on certain objects. We refer to our model as Uncertainty-Aware Graph network for object DETection (UAGDet). We then experimentally validate ours on the challenging large-scale aerial image dataset, namely DOTA, that consists of lots of objects with small to large sizes in an image, on which ours improves the performance of the existing object detection network.


\end{abstract}
\section{Introduction}
Given an input image, the goal of object detection is to find the bounding boxes and their corresponding classes for objects of interests in the image. To tackle this task, various object detection models based on conventional convolutional neural networks (CNNs), including Faster R-CNN~\cite{FasterRCNN} and YOLO~\cite{YOLO}, to recent transformer-based~\cite{Transformers} models, such as DETR~\cite{DETR} and Deformable DETR~\cite{deformabletransformer}, are proposed, showing remarkable performances. In other words, there have been considerable attentions to search for new architectures, for improving their performances on various object detection tasks, for example, finding objects in aerial images from Earth Vision~\cite{DOTA}.

However, despite their huge successes, most existing models are limited in that they do not consider interactions among objects explicitly, which are different from how humans perceive images: considering \textit{how each object is spatially and semantically related to every other objects}, as well as capturing visual features of local regions, for contextualizing the given image. Thus, we suppose that a scheme that does model explicit relationships among objects in the image is necessary for any object detection networks. Note that it becomes more important when handling aerial images~\cite{DOTA}, which we mainly target, that consist of a large number of objects with varying scales from extremely small to large than images in the conventional datasets (i.e., COCO~\cite{COCO} or Pascal VOC~\cite{PascalVOC}). 
\begin{figure*}[t]
    \begin{minipage}{1\textwidth}
    \centering
    \includegraphics[width=1.0\linewidth]{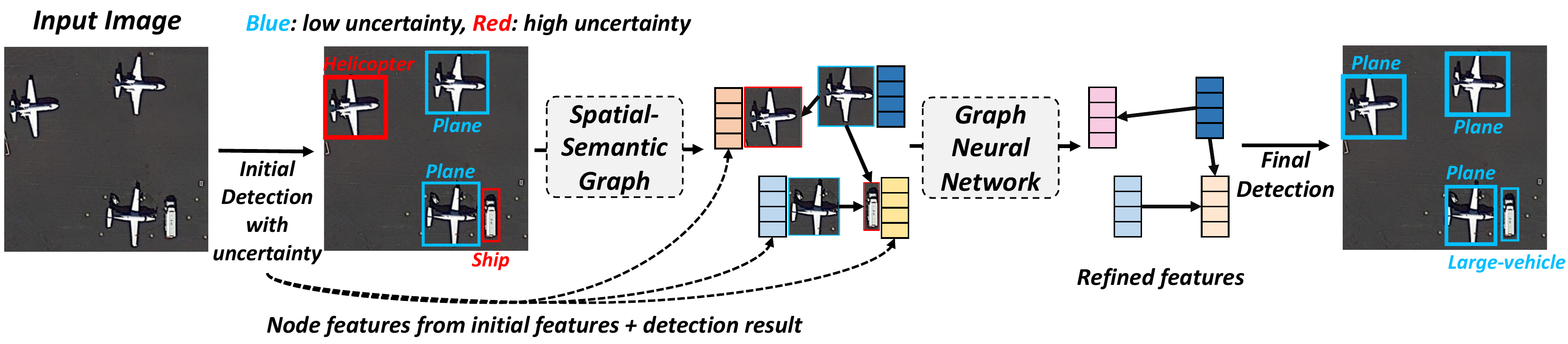}
    \end{minipage}
    \caption{\textbf{Concept Figure.} Given an input image, initial object detection is done, which generates uncertainties of objects as well. \textcolor{cyan}{Blue} objects with low uncertainties and \textcolor{red}{red} objects with high uncertainties are set as \textit{source} and \textit{target} nodes, respectively. A directed graph connecting source nodes to target nodes, where edges are generated based on spatial-semantic distances between nodes, is fed into GNNs to obtain refined representations of objects.}
    \label{fig:concept}
\end{figure*}
The most straightforward approach to consider relationships in the image is to construct an inherent hierarchy, where local-level features for identifying local parts are obtained by outputs of CNNs on sub-regions, whereas global-level features for overall understanding of an entire image are obtained by aggregating locally obtained CNN features~\cite{SSD,FPN,dai2022tardet}. However, this scheme is highly suboptimal, since it does not explicitly consider semantic and spatial relationships among objects with object representations, but it does implicitly model their relationships with CNN features from different sub-regions. Therefore, in this work, we aim to first detect objects in the given input image with conventional object detection networks (e.g., Faster R-CNN~\cite{FasterRCNN} or RoITransformer~\cite{RoITransformer}), and then calculate the edges between them with their distances of semantic and spatial features. After that, over the constructed graph with objects as nodes and their distances as edges, we leverage the graph neural networks (GNNs)~\cite{GCN,GNN} to obtain graphical features of objects, which are then combined with previously obtained visual CNN features for improving final object prediction performances. Note that, recently, there exist few similar works~\cite{relationnetwork,SGRN} that consider explicit interactions between all objects by constructing either a fully connected object graph~\cite{relationnetwork} or a spatial-aware sparse graph~\cite{SGRN}, whose object representations are used to improve CNN features for object detection. However, is it necessary to refine CNN features and detection results of every object?

We suppose that, if the prediction results of object detection networks are sufficiently certain, it would not be necessary to replace their features and prediction labels. Therefore, in this work, we aim to improve the prediction results only on uncertain objects that detection models are mostly confused about, by leveraging the \textit{uncertainty} on each object when constructing object graphs and refining object features in the given image. Specifically, when we construct a graph composed by objects (nodes) and their both semantic and spatial similarities (edges), we propose to propagate the information from certain objects to uncertain ones by defining directions on edges represented by a directed graph. This scheme allows the model to improve the features of uncertain objects by contextualizing them with their semantically- and spatially-related certain ones, but also prevents the certain objects to receive noisy information from objects that are uncertain, which is also highly efficient because of ignoring a large number of edges between certain objects compared against existing models~\cite{relationnetwork,SGRN} that consider all object pairs. Moreover, when we refine CNN features by representations of the object graph from GNNs, we propose to manipulate the features for only uncertain objects, rather than changing all object features, which allows the model to focus on improving uncertain objects while maintaining the high prediction performances on certain objects. Lastly, to further make the model focus on uncertain objects, we scale the training loss based on uncertainties, where uncertain predictions get higher weights in their losses.

We refer our novel object detection framework, which explicitly models relationships between objects over the directed graph with their uncertainties, as Uncertainty-Aware Graph network for object DETection (UAGDet), which is illustrated in Fig.~\ref{fig:concept}. We then experimentally validate our UAGDet on the large-scale Dataset for Object deTection in Aerial images (DOTA)~\cite{DOTA} with two different versions: DOTA-v1.0 and DOTA-v1.5, containing large numbers of small and uncertain objects with sufficient interaction existing among them. Therefore, we suppose our UAGDet is highly beneficial in such a challenging scenario. Experimentally, we use RoITrans~\cite{RoITransformer} as a backbone object detection network, and the results show that our UAGDet shows 2.32\% and 3.82\% performance improvements against the backbone RoITrans network on DOTA-v1.0 and -v1.5 datasets, respectively, with mean Average Precision (mAP) as an evaluation metric. Also, further analyses show that our graph construction method considering both semantic and spatial features connects related objects by edges, and also uncertainties are measured high on incorrect predictions which are corrected by refined features from graph representations. 
\section{Related Works}

\subsubsection{Object Detection.}
Object detection is the task of localizing and classifying objects of interest in an image. To mention a few, Faster R-CNN~\cite{FasterRCNN} and YOLO~\cite{YOLO} are two CNN-based early models, successfully demonstrating deep learning's capability in the object detection task. Recently, following the success of transformers~\cite{Transformers} in natural language processing, DETR~\cite{DETR} adopts transformers in the object detection task, which performs self-attention on CNN features from different sub-regions of an input image to consider their relationships. Furthermore, Deformable DETR~\cite{deformabletransformer} is proposed to attend only to the relative spatial locations when performing self-attention rather than considering all locations, which makes the transformer-based prediction architecture efficient. However, despite their huge improvements in regard to developing new object detection networks and their performances, they largely overlook the relationships among objects. In particular, CNN-based models rely on constructing hierarchical structures of local and global features for capturing feature-level relationships via multi-level feature map, which is known as a feature pyramid network (FPN)~\cite{FPN}, but not capturing the explicit object-level relationships. Meanwhile, transformer-based models only implicitly consider relationships between candidate queries for objects, but also candidate queries often include duplicated objects or meaningless background areas which particularly belong to 'no object'. Thus, unlike these previous works, we propose to explicitly model object-level relationships, by initially detecting objects and then, over the graph structure with objects as nodes, representing them with graph neural networks (GNNs) for obtaining graphical features, which are then combined with CNN features for final object detection.  

\subsubsection{Graph Neural Networks for Object Detection.}
Graph Neural Networks (GNNs), which expressively represent graph structured data consisting of nodes and edges by iteratively aggregating features from target node's neighbors, have gained substantial attention in recent years, showing successes in various downstream tasks working on graphs~\cite{GCN,graphsage,GAT,GATv2,jo2021edge,GNN}. As such model architectures explicitly leverage the relationships between connected instances when representing them, there have been recent attempts to use GNNs on the object detection task to capture interactions among objects~\cite{SGRN,GAR,RelationRCNN}. For instance, GAR~\cite{GAR} constructs a context graph, which considers interactions between objects and scenes but also between the objects themselves by forming objects and scenes as nodes and their connections as edges, and then represents the graph with GNNs. Similarly, Relation R-CNN~\cite{RelationRCNN} generates semantic and spatial relation networks with objects as nodes, using pre-built co-occurrence matrix and distances between objects, respectively, for forming edges. However, both GAR~\cite{GAR} and Relation R-CNN~\cite{RelationRCNN} have an obvious limitation that the edge generation procedure is not end-to-end trainable, but based on simple heuristics using pre-calculated statistics of co-occurrences between instances. On the other hand, SGRN~\cite{SGRN} learns a spatial-aware relation network between objects based on visual features and spatial distances of initial proposals, which is then forwarded to GNNs for obtaining representations for objects. Note that our work has key differences against such a relevant work: instead of working on all object proposals, we first largely reduce them with non-maximum suppression (NMS)~\cite{dalal2005nms1,felzenszwalb2010nms2}, and then aim at improving only the uncertain objects that the model is largely confused about by generating semantic and spatial yet directed edges from certain objects to uncertain ones, which greatly reduces computational costs especially when dealing with lots of objects in a single image (e.g., DOTA~\cite{DOTA}).

\subsubsection{Uncertainty-Aware Object Detection.}
One of the main focuses of Bayesian deep learning in computer vision tasks is to accurately capture uncertainty of a model's prediction, which allows the deep learning models to prevent making incorrect predictions when their uncertainties are high. Following the successful applications of uncertainty on semantic segmentation and depth estimation tasks~\cite{Gal2016Uncertainty,WhatUncertainties,UncertDistill}, some works propose to utilize uncertainty in the object detection task as well~\cite{varvote,CertainNet}. In particular, Xu et al.~\cite{varvote} propose to merge overlapping bounding boxes based on the weights of their coordinates obtained from uncertainties (i.e., the larger the uncertainties, the lower the weights) when performing soft-NMS~\cite{SoftNMS}, which contributes to accurately localizing bounding boxes. Also, CertainNet~\cite{CertainNet} measures uncertainties using the distances between predicted features and learnable class centroids (i.e., the larger the distances at inference time, the higher the uncertainties), which is based on the work of deterministic uncertainty quantification (DUQ)~\cite{van2020uncertainty}. However, it does not use uncertainty for improving model's prediction, but only calculates uncertainty values of predicted objects. Note that, in contrast to those previous works~\cite{varvote,CertainNet} that use uncertainty either on combining overlapping bounding boxes or on identifying less certain objects, we leverage uncertainties in totally different perspectives. That is we propose to refine representations of uncertain objects by transferring knowledge from certain objects to uncertain ones over the directed graph structure, which may correct initially misclassified objects with high uncertainties.
\section{Method}

\subsubsection{Faster R-CNN Baseline.}
We first define the notations and terms for objection detection by introducing the object detection pipeline of the Faster R-CNN model~\cite{FasterRCNN}. In Faster R-CNN, initial $N$ proposals $prop_i (i = 1, ..., N)$ are generated from an input image by Region Proposal Networks (RPN). Then, we obtain $N$ CNN feature maps $\hat{f}^{conv}_i$, each of which is associated with its proposal $prop_i$, with Region of Interest (RoI) extraction modules (e.g., RoIPool~\cite{RCNN} or RoIAlign~\cite{MaskRCNN}), to make use of such features for finding bounding boxes and their classes. Specifically, extracted CNN features are fed into two separated fully connected layers, called classification head and regression head, to obtain the class prediction result $p_i$ and the coordinates of detected bounding box $bbox_i$ for each proposal, respectively. The entire process is formally defined as follows:
\begin{equation}
\begin{split}
    & prop_i, conv\_feature = \texttt{RPN}(image), \\
    & \hat{f}^{conv}_i = \texttt{RoIAlign}(prop_i, conv\_feature), \\
    & \widehat{p_i} = \texttt{cls\_head}(\hat{f}^{conv}_i), \; \widehat{bbox_i} = \texttt{reg\_head}(\hat{f}^{conv}_i),
\end{split}
\label{eq:rcnnpipeline}
\end{equation}
where $\texttt{RPN}$ is a region proposal network for extracting object proposals, $\texttt{RoIAlign}$ is a CNN feature extraction module for each proposal, $\texttt{cls\_head}$ is a object classification network, and $\texttt{reg\_head}$ is a bounding box regression network. $\widehat{p_i}$ and $\widehat{bbox_i}$ denote predicted class label and bounding box coordinates, respectively.

To optimize a model during training, classification loss $L_{cls}$ is calculated from classification results with \texttt{CrossEntropy} loss, and regression loss $L_{reg}$ is calculated from regression results with \texttt{SmoothL1} loss~\cite{FastRCNN}, defined as follows:
\begin{equation}
    L_{cls} = \sum^{N}_{i}\texttt{CrossEntropy}(\widehat{p_i}, y_i), \; L_{reg} = \sum^{N}_{i}\texttt{SmoothL1}(\widehat{bbox_i}, bbox_i),
\end{equation}
where $y_i$ denotes an one-hot vector of a ground-truth label of a proposal $i$, and $bbox_i$ denotes a ground-truth bounding box coordinates. Then, based on two object detection losses $L_{cls}$ and $L_{reg}$, overall training loss is formulated as a weighted sum of them: $L_{obj} = L_{cls} + \lambda_1 \times L_{reg}$, where $\lambda$ is a scaling factor. Note that, in the test time, Non-Maximum Suppression (NMS) is applied so that duplicated bounding boxes with high overlapping areas and non-promising bounding boxes with low confidences are removed. 

\subsubsection{Overview of Our Uncertainty-Aware Graph Network.}
In this work, we aim to further improve the object detection network, for example, Faster R-CNN, by leveraging the object representations over the directed graph, which transfers knowledge from certain objects to uncertain objects by reflecting their semantic-spatial distances via Graph Neural Networks (GNNs). To achieve this goal, we first measure the uncertainty of every detected object with minimal costs, while following the initial object detection pipeline represented in equation~\ref{eq:rcnnpipeline}, which we specify in Subsection~\ref{sec:lightMCDropout}. Then, we construct an object graph consisting of objects as nodes and their spatial and semantic relatedness as edges, to contextualize all objects in the input image with a message passing scheme of GNNs. However, since changing features and prediction results for certain objects may not be necessary, we focus on improving uncertain objects by leveraging their contextual knowledge with semantic-spatial relationships to certain objects. We introduce this uncertainty-based graph generation procedure in Subsection~\ref{sec:graphconstruction}. After that, based on object representations obtained by GNNs over the object graph: node features from visual CNN features and initial object classes; edge weights from pairwise spatial distances among objects, we refine the features for uncertain objects to improve their prediction performances, which is described in Subsection~\ref{sec:featurerefine}. We finally summarize our overall object detection pipeline in Subsection~\ref{sec:pipeline}, which is also illustrated in Figure~\ref{fig:concept}. 

\subsection{Uncertainty-Aware Initial Object Detection}
\label{sec:lightMCDropout}
Using the Faster R-CNN pipeline defined above, we can predict $N$ classification results and their corresponding bounding boxes. Then, in this subsection, we describe how to measure the uncertainty of detected objects along with the Faster R-CNN architecture, to find out the target objects worth refining.

We first define the uncertainty of each object as $\phi_i$. Then, to measure this, we use MC Dropout~\cite{MCDropout}. In particular, MC dropout can approximate an uncertainty of outputs, obtained from $M$ times of repeated forward passes while enabling dropout, by calculating a variance of them. To alleviate the excessive computational cost of MC Dropout caused by repeated forward passes through every layer, we introduce a slightly modified version of the original MC dropout. This approach is simple -- instead of repeating whole forward passes, we only repeat forward passes with dropout in last fully connected layers of Faster R-CNN model's classification head and regression head, that are $\texttt{cls\_head}$ and $\texttt{reg\_head}$ in equation~\ref{eq:rcnnpipeline}. Therefore, we can measure each object's uncertainty with minimal additional computational costs. Also, the suggested uncertainty measure can be easily implemented to other object detection frameworks, since the only modification required is simply repeating last forward steps $M$ times with dropout. The formal algorithm of the lightweight MC Dropout is shown in Algorithm \ref{alg:LightMCDropout}.

\setlength{\textfloatsep}{10pt}
\begin{algorithm}[t!]
\caption{Lightweight MC Dropout}\label{alg:opt}
\label{alg:LightMCDropout}
\textbf{Input:} an image $I$, and the number of dropout iterations $M$. \\
\textbf{Outputs:} $N$ numbers of classification results $\{p_1, ..., p_N\}$ with their corresponding bounding boxes $\{bbox_1, ..., bbox_N\}$ and uncertainties $\{\phi_1, ..., \phi_N\}$, where $N$ denotes the number of object proposals.
\renewcommand{\algorithmiccomment}[1]{//#1}
\begin{algorithmic}[1]
    \STATE $\{(prop_i, conv\_feature)\}_{i=1}^{N} \leftarrow \texttt{RPN}(I)$
    \STATE $\{(\hat{f}^{conv}_i)\}_{i=1}^{N} \leftarrow \{\texttt{RoIAlign}(prop_i, conv\_feature)\}_{i=1}^N$
   \FOR{$j \leftarrow 1, \ldots, M$}
   \STATE $\{p^{j}_i\}_{i=1}^N \leftarrow \{\texttt{cls\_head}(\hat{f}^{conv}_i)\}_{i=1}^N$
   \STATE $\{bbox^{j}_i\}_{i=1}^N \leftarrow \{\texttt{reg\_head}(\hat{f}^{conv}_i)\}_{i=1}^N$ 
   \ENDFOR \hfill\COMMENT{$p$ and $bbox$ are stacked to shape : $(N, M, ...)$}
   \STATE $p_i \leftarrow \texttt{mean}(\{p^j_i\}_{j=1}^M)$ 
   \STATE $bbox_i \leftarrow \texttt{mean}(\{bbox^j_i\}_{j=1}^M)$ 
   \STATE $\phi_i \leftarrow \texttt{stdev}(\{p^{j}_i\}_{j=1}^M)$
   \STATE \textbf{return} $\{p_1, ..., p_n\}, \{bbox_1, ..., bbox_n\}, \{\phi_1, ..., \phi_n\}$
\end{algorithmic}
\end{algorithm}

After the object detection step above, we apply NMS with 1/10 of the usual threshold value to remove redundant and meaningless objects while maintaining source and targets objects, instead of excessively removing lots of objects except for the most certain ones. Consequently, after the initial detection and NMS phases, $N_{nms}$ objects with their features and uncertainties remain. Note that when any vector existing at this point is used in subsequent layers, their gradients are detached so that the initial detection model is left unaffected by the following modules.

\subsection{Uncertainty-Based Spatial-Semantic Graph Generation}
\label{sec:graphconstruction}
In this subsection, we now explain how to construct an object graph based on objects' uncertainties and their spatial-semantic distances. At first, we first sort all $N_{nms}$ objects based on their uncertainties in ascending order. Then, the top half with low uncertainties belongs to the source set $\mathcal{V}_{src}$, and the bottom half with high uncertainties belongs to the target set $\mathcal{V}_{dst}$. After categorizing all objects into source and target, we construct a directed bipartite graph $\mathcal{G}$ that consists of a set of object nodes $\mathcal{V} = \{\mathcal{V}_{src}\cup \mathcal{V}_{dst}\}$ and their edge set $\mathcal{E}$, where source node $v_{src}$ and target node $v_{dst}$ belong to source set $\mathcal{V}_{src}$ and target set $\mathcal{V}_{dst}$, respectively: $v_{src} \in \mathcal{V}_{src}$ and $v_{dst} \in \mathcal{V}_{dst}$. Note that, in our directed bipartite graph, every edge $e_i \in \mathcal{E}$ connects only the source node to the relevant target node but not in the reverse direction, so that only target nodes (i.e., uncertain nodes) are affected by source nodes (i.e., certain nodes). 

Then, to share the knowledge only across relevant nodes, instead of connecting every pair of nodes, we aim at connecting a \textit{related} pair of nodes. To do so, based on the motivation that each object is likely to be affected by its nearby objects, our first criterion of relatedness is defined by a spatial distance between source and target nodes.
Specifically, we calculate the Euclidean distance between two nodes' coordinates $c_{src}$ and $c_{dst}$ and consider it as a spatial distance measure, which is formally defined as follows: $d^{spa}_{(v_{src}, v_{dst})} = ||c_{src} - c_{dst} ||^2$. Here, if $d^{spa}_{(v_{src}, v_{dst})}$ is smaller than the certain spatial threshold value $thr_{spa}$, we add an edge $e_i = (v_{src}, v_{dst})$ to the graph. Therefore, every target node has edges to its nearby source nodes. 

On the other hand, we also consider the semantic distance between object nodes to further contextualize them based on their representation-level similarities. To be specific, we similarly calculate the Euclidean distance between CNN features of source and target objects, which is considered as a semantic distance and formally defined as follows: $d^{sem}_{(v_{src}, v_{dst})} = ||\hat{f}^{conv}_{src} - \hat{f}^{conv}_{dst} ||^2$, where $\hat{f}^{conv}_{src}$ and $\hat{f}^{conv}_{dst} $ denote CNN feature maps for source and target objects, respectively. After that, similarly in the edge addition scheme for spatial distances, if $d^{sem}_{(v_{src}, v_{dst})}$ is smaller than the semantic threshold $thr_{sem}$, we add an edge $e_i = (v_{src}, v_{dst})$.

\subsection{Feature Refinement via GNNs with Spatial-Semantic Graph}
\label{sec:featurerefine}
With the spatial-semantic graph $\mathcal{G} = (\mathcal{V}, \mathcal{E}$) built above, we aim at refining the target (i.e., uncertain) node representations by aggregating features from their source (i.e., certain) node neighbors via GNNs. Before doing so, we have to define node features and edge weights, which we describe in the following paragraphs. 

At first, we aim to initialize the node features as features of the CNN outputs and the predicted class. To do so, we first apply 1$\times$1 convolutions to CNN features for each object, to generate a more compact visual representation $\hat{f}^{down}_i$ by downsizing original features $\hat{f}^{conv}_i$ in equation~\ref{eq:rcnnpipeline}, therefore having half of the original channel dimension. Also, to explicitly make use of the predicted class information, we regard the index of the maximum value (e.g., argmax) in the initial probability vector $\widehat{p_i}$ as its class, and then embed it into the representation space. After that, by concatenating both the down-sized CNN features and the class embedding vector, we initialize the node features $\hat{f}^{node}_i$ for each node. The overall process is formally represented in equation~\ref{eq:GNNFeatureGeneration} below.  

On the other hand, edge weight used for neighborhood aggregation in GNNs is defined as a reciprocal of a pairwise spatial distance between source and target nodes divided by the diagonal length of the input image $I$ for normalization, formulated as follows: $w_{i} = 1 / (d^{spa}_{(v_{src}, v_{dst})} \times \texttt{diag\_len}(I))$ for $e_{i} = (v_{src}, v_{dst})$, where $\texttt{diag\_len}$ denotes a function returning the length of a diagonal of the given image. Note that such an edge weighting scheme allows the GNN model to give larger weights on nearby objects during the aggregation of features from neighboring source nodes to the target node. For GNN, we use two Graph Convolutional Network (GCN)~\cite{GCN} layers, and the overall graph feature extraction procedure is as follows:
\begin{equation}
\begin{split}
    & \hat{f}^{down}_i = \texttt{1x1Conv}(\hat{f}^{conv}_i), \; \hat{c}_i = \texttt{embed}(\texttt{argmax}(\hat{p}_i)) \\
    & \hat{f}^{node}_i = \texttt{concat}(\hat{f}^{down}_i, \hat{c}_i), \; w_{i} =  1 / (d^{spa}_{(v_{src}, v_{dst})} \times \texttt{diag\_len(I))}, \\
    & \hat{f}^{gnn}_i = \texttt{GCN}(\mathcal{E}, \hat{f}^{node}, w),
\end{split}
\label{eq:GNNFeatureGeneration}
\end{equation}
where $\texttt{1x1Conv}$ denotes the 1$\times$1 convolutional operation, $\texttt{argmax}$ returns the index of the maximum value, $\texttt{embed}$ denotes the class embedding function, $\texttt{concat}$ denotes the concatenation operation, and we use the GCN for representing nodes in \texttt{GCN}. Based on equation~\ref{eq:GNNFeatureGeneration}, the resulting node features for each object $\hat{f}^{gnn}_i$ capture visual and class features based on its relationships to other certain objects as well as itself, but also capture explicit relatedness between source and target nodes via their spatial distances.

\subsection{Final Detection Pipeline and Training Losses}
\label{sec:pipeline}
In this subsection, we describe the final detection process, which is done with contextualized object representations from GNNs. To be specific, for each object, we first concatenate the visual features $\hat{f}^{conv}_i$ from CNN layers and the graphical features $\hat{f}^{gnn}_i$ from GNN layers in a channel-wise manner. Then, we forward the concatenated features $\hat{f}^*_i = \texttt{concat}(\hat{f}^{conv}_i, \hat{f}^{gnn}_i)$ to the fully-connected layers to obtain the final class probability vector $\hat{p}^*_i$, which is similar to the class prediction process in equation~\ref{eq:rcnnpipeline} for Faster R-CNN, while we use the differently parameterized class prediction head $\texttt{cls\_head}^*$. That is defined as follows:
\begin{equation}
\begin{split}
    & \hat{f}^*_i = \texttt{concat}(\hat{f}^{conv}_i, \hat{f}^{gnn}_i), \\
    & \hat{p}^{*}_i = \texttt{cls\_head}^*(\hat{f}^*_i). \\
\end{split}
\label{eq:ourpipeline}
\end{equation}

Note that one of the ultimate goals of our method is to improve the performance on uncertain objects while maintaining high performance on certain objects. Therefore, we additionally regulate the \texttt{CrossEntropy} loss based on the input prediction's uncertainty value: we give larger weights to objects with high uncertainties so that the model could focus on those objects. To do so, we first use the \texttt{softmax} function to normalize the weight values, and then multiply the number of target objects $N_{nms}$ to keep the scale of the loss: the sum of all $w_i$ from $i=1$ to $N$ equals to the number of objects $N_{nms}$, as follows:
\begin{equation}
\begin{split}
    w_i &= \texttt{softmax}(\phi_i) \times N_{nms}, \\
    &= \frac{\texttt{exp}(\phi_i/\tau)}{\sum^{N_{nms}}_{j=1} \texttt{exp}(\phi_j/\tau)} \times N_{nms}, \\
\end{split}
\label{eq:uncertaintyweight}
\end{equation}
where $\tau$ denotes the temperature scaling value for weights. Then, based on the loss weight for each object proposal, classification loss $L_{ref}$ from refined features consisting of CNN and GNN representations is defined as follows:
\begin{equation}
\begin{split}
    & L_{ref} = \sum^{N_{nms}}_{i} w_i \times \texttt{CrossEntropy}(\hat{p}^{*}_i, y_i). \\
\end{split}
\label{eq:graphloss}
\end{equation}

The overall training loss is then defined by the initial and refined object detection losses, $L_{obj}$ and $L_{ref}$, as follows: $L_{total} = L_{obj} + \lambda_2 \times L_{ref}$, where $\lambda_2$ is the scaling term for the last loss. Note that we refer our overall architecture calculating $L_{ref}$ as Uncertainty-Aware Graph network for object DETection (UAGDet), which is jointly trainable with existing object detection layers by the final objective $L_{total}$, and easily applicable to any object detection networks. In the test time, we apply NMS to the final object detection results to remain only the most certain bounding boxes and their classes. Also, during evaluation, we replace the initial detection results for uncertain objects associated with target nodes with their final object detection results, while maintaining the initial detection results for certain objects, to prevent the possible performance drop.

\section{Experiments}
In this section, we validate the proposed Uncertainty-Aware Graph network for object DETection (UAGDet) for its object detection performance on the large-scale Dataset for Object deTection in Aerial images (DOTA)~\cite{DOTA}. 

\subsection{Datasets}
DOTA~\cite{DOTA} is widely known as an object detection dataset in aerial images. There are three versions of DOTA datasets, and we use two of them for evaluation: DOTA-v1.0, and v1.5. Regarding the dataset statistics, DOTA-v1.0 contains 2,806 large-size images from the aerial view, which are then processed to have 188,282 object instances in total within 15 object categories. The 15 object categories for classification are as follows: Plane (PL), Baseball-diamond (BD), Bridge (BR), Ground track field (GTF), Small vehicle (SV), Large vehicle (LV), Ship (SH), Tennis court (TC), Basketball court (BC), Storage tank (ST), Soccer ball field (SBF), Roundabout (RA), Harbor (HB), Swimming pool (SP), and Helicopter (HC). The number of instances per class is provided in Table~\ref{tab:result}. DOTA-v1.5 uses the same images as in DOTA-v1.0, while extremely small-sized objects and a new category named Container Crane (CC) are additionally annotated. Therefore, DOTA-v1.5 contains 403,318 object instances with 16 classes. 

Note that the bounding box regression of the DOTA dataset is different from the conventional datasets, which makes object detection models more challenging. Specifically, in conventional object detection datasets, such as COCO~\cite{COCO} or Pascal VOC~\cite{PascalVOC}, the ground truth bounding box of each object is annotated as $(x, y, w, h)$ format. However, since objects have a wide variety of orientations in aerial images, the ground truth bounding box in the DOTA~\cite{DOTA} dataset is denoted as $(x, y, w, h, \theta)$ format, with an additional \textit{rotation angle} parameter $\theta$. To reflect such an angle in the object detection architecture, RoITransformer~\cite{RoITransformer} uses additional fully-connected layers to regress additional parameter $\theta$ on top of the Faster R-CNN architecture~\cite{FasterRCNN}, which we follow in our experiments. For model tuning and evaluation on the DOTA dataset, we follow the conventional evaluation setups~\cite{RoITransformer,DOTA,DOTA2}. We first tune the hyperparameter of our UAGDet on the $val$ dataset while training the model on the $train$ set. After the tuning is done, we use both the $train/val$ sets to train the model, and infer on the $test$ set. Inference results on the $test$ set is uploaded to the DOTA evaluation server~\cite{DOTA,DOTA2}, to measure the final performance. For evaluation metrics, we use the Average Precision for each class result, and also the mAP for all results~\cite{DOTA}.

\subsection{Experimental Setups}

\subsubsection{Baselines and Our Model.}
We compare our UAGDet with Mask R-CNN~\cite{MaskRCNN,DOTA2}, RoITransformer~\cite{RoITransformer} and GFNet~\cite{GFNet} models. Specifically, Mask R-CNN~\cite{MaskRCNN} is applied to DOTA~\cite{DOTA} by viewing bounding box annotations as coarse pixel-level annotations and finding minimum bounding boxes based on pixel-level segmentation results~\cite{DOTA2}.
RoITransformer (RoITrans)~\cite{RoITransformer} is based on Faster R-CNN~\cite{FasterRCNN} with one additional FC layers in order to predict rotation parameter $\theta$ for rotated bounding boxes. GFNet~\cite{GFNet} uses GNNs to effectively merge dense bounding boxes with a cluster structure, instead of using algorithmic methods such as NMS or Soft-NMS~\cite{SoftNMS}. Our UAGDet uses GNNs for object graphs composed by objects' uncertainties to improve uncertain object predictions with graph representations, where we use RoITrans as a base object detection network.

\subsubsection{Implementation Details.} 
Regarding the modules for the architecture including CNN backbone, RPN, and regression/classification heads, we follow the setting of RoITrans~\cite{RoITransformer}. The number of region proposals is set to 1,250. Based on the objects' uncertainties, we set top half nodes as source nodes while bottom half nodes as target nodes. If $N_{nms}$ is larger then 100, we only consider top 100 objects, and exclude the rest of highly uncertain objects that are likely to be noise. We set the dropout ratio as 0.2, $M$ for MC dropout as 50, $\lambda_1, \lambda_2$ for loss weights as 1, $thr_{spa}$ for sparse graph generation as 50, and $thr_{sem}$ for semantic graph generation as 10. The dimensionality of tensors in equation~\ref{eq:GNNFeatureGeneration} is as follows: $\hat{f}^{conv}_i \in \mathbb{R}^{256 \times 7 \times 7}$, $\hat{f}^{down}_i \in \mathbb{R}^{128 \times 7 \times 7}$, and $\hat{c}_i \in \mathbb{R}^{16 \times 7 \times 7}$. For GNN layers, we use two GCN layers~\cite{GCN} with \texttt{LeakyReLU} activation between them. Concatenated feature $f^{node}_i$ for each node is first fed into \texttt{1x1Conv} for reducing its dimension from $\mathbb{R}^{144 \times 7 \times 7}$ to $\mathbb{R}^{128 \times 7 \times 7}$, and then fed into \texttt{GCN}, where its dimension is further reduced by 1/2 and 1/4 in two GCN layers, respectively, i.e., $\hat{f}^{gnn}_i \in \mathbb{R}^{16 \times 7 \times 7}$. We set a temperature scale value $\tau$ in $\texttt{softmax}$ of equation~\ref{eq:uncertaintyweight} as 0.1. The model is trained for 12 epochs with a batch size of 4, a learning rate of 0.01, and a weight decay of $10^{-4}$, and optimized by a SGD. We use 4 TITAN Xp GPUs.

\subsection{Quantitative Results}
\subsubsection{Main Results.}
\begin{table*}[t]
    
    \caption{Object detection results on DOTA-v1.0 and DOTA-v1.5 test datasets. The CC (Container Crane) category only exists in DOTA-v1.5, thus we remain the performance of it on DOTA-v1.0 as blank. Best results are marked in bold.}
    \begin{adjustbox}{width=\textwidth}
    \renewcommand{\arraystretch}{0.95}
    \begin{tabular}{ll  ccccccccccccccccc}
    \toprule
    & Models & PL & BD & BR & GTF & SV & LV & SH & TC & BC & ST & SBF & RA & HB & SP & HC & CC & \textbf{mAP} \\
    \toprule
    
    \multirow{6}{*}{\rotatebox{90}{\textit{\fontsize{8pt}{8pt}\selectfont v1.0}}}
    & \# of instances & 14,085 & 1,130 & 3,760 & 678 & 48,891 & 31,613 & 52,516 & 4,654 & 954 & 11,794 & 720 & 871 & 12,287 & 3,507 & 822 & - \\ \cdashline{2-19}\noalign{\vskip 0.75ex}
    & Mask R-CNN & 88.7 & 74.1 & 50.8 & 63.7 & 73.6 & 74.0 & 83.7 & 89.7 & 78.9 & 80.3 & 47.4 & 65.1 & 64.8 & 66.1 & 59.8 & - & 70.7\\
    & GFNet & 90.3 & 83.3 & 51.9 & 77.1 & 65.5 & 58.2 & 61.5 & 90.7 & 82.1 & 86.1 & 65.3 & 63.9 & 70.6 & 69.5 & 57.7 & - & 71.6\\
    & RoITrans & 87.9 & 81.1 & 52.9 & 68.7 & 73.9 & 77.4 & 86.6 & 90.2 & 83.3 & 78.1 & 53.5 & 67.9 & 76.0 & 68.7 & 54.9 & - & 73.4\\
    & UAGDet (Ours) & 89.3 & 83.3 & 55.5 & 73.9 & 68.4 & 80.2 & 87.8 & 90.8 & 84.2 & 81.2 & 54.3 & 61.8 & 76.8 & 70.9 & 68.0 & - & \textbf{75.1} \\
    \midrule
    
    \multirow{5}{*}{\rotatebox{90}{\textit{\fontsize{8pt}{8pt}\selectfont v1.5}}}
    & \# of instances & 14,978 & 1,127 & 3,804 & 689 & 242,276 & 39,249 & 62,258 & 4,716 & 988 & 12,249 & 727 & 929 & 12,377 & 4,652 & 833 & 237 \\ \cdashline{2-19}\noalign{\vskip 0.75ex}
    & Mask R-CNN & 76.8 & 73.5 & 50.0 & 57.8 & 51.3 & 71.3 & 79.8 & 90.5 & 74.2 & 66.0 & 46.2 & 70.6 & 63.1 & 64.5 & 57.8 & 9.42 & 62.7\\
    & RoITrans & 71.7 & 82.7 & 53.0 & 71.5 & 51.3 & 74.6 & 80.6 & 90.4 & 78.0 & 68.3 & 53.1 & 73.4 & 73.9 & 65.6 & 56.9 & 3.00 & 65.5\\
    & UAGDet (Ours) & 78.4 & 82.4 & 54.4 & 74.1 & 50.7 & 74.2 & 81.0 & 90.9 & 79.3 & 67.0 & 52.3 & 72.8 & 75.8 & 72.4 & 65.3 & 15.4 & \textbf{68.0} \\
    \bottomrule
    
    \end{tabular}
    \end{adjustbox}
    \label{tab:result}
\end{table*}

We report the performances of baseline and our models on both DOTA-v1.0 and DOTA-v1.5 datasets in Table \ref{tab:result}. As shown in Table~\ref{tab:result}, our UAGDet largely outperforms all baselines on both datasets in terms of mAP, obtaining 1.7 and 2.5 point performance gains on DOTA-v1.0 and v1.5 datasets, respectively against RoITrans baseline. The difference between the baseline model and ours is more dramatic in the DOTA-v1.5 dataset, which matches our assumption: our UAGDet is more beneficial in DOTA-v1.5 containing extremely small objects since they get effective representations based on interaction information. Furthermore, our model outperforms GFNet~\cite{GFNet}, which first builds instance clusters and then applies GNNs to learn comprehensive features of each cluster, since we do not impose a strong assumption as in GFNet~\cite{GFNet} that cluster always exists in an image, and also we consider uncertainties of objects in the graph construction and representation learning scheme. Note that our UAGDet only uses 1,250 proposals instead of 2,000 for computational efficiency and no augmentation is applied, thus we believe further performance gains could be easily achieved by additional computation if needed, for real-world applications.

\subsubsection{Ablation Study.}
To see where the performance gain comes from, we conduct an ablation study on the DOTA-v1.0 dataset, and report the results in Table \ref{tab:ablation}. In particular, we ablate two components of our UAGDet: node and edge
\begin{wraptable}{r}{0.3\linewidth}
\small
\centering
\caption{\small \textbf{Results of an ablation study} on the DOTA-v1.0 dataset.}
\label{tab:ablation}
\begin{adjustbox}{width=1.\linewidth}
\begin{tabular}{@{}l c@{}}
\toprule
\small
Models & mAP \\
\midrule \midrule
\textbf{UAGDet (Ours)} &  \textbf{75.1} \\
w/o Complex Feature    &  73.9 \\
w/o Uncertainty-scaled Loss &  74.7  \\
\bottomrule
\end{tabular}
\end{adjustbox}

\end{wraptable}
features in GNNs; uncertainty-scaled loss in equation~\ref{eq:uncertaintyweight}. At first, we only use the CNN feature maps as node features, instead of using initially predicted class embedding for nodes and pairwise spatial distances for edges, and we observe the large performance drop of 1.2 point.
Furthermore, we do not use the uncertainty-scaled loss for training but rather use the naive \texttt{CrossEntropy} loss, and we observe the performance drop of 0.4 point. Those two results suggest that, using complex features in GNN layers, as well as applying uncertainty-aware losses for focusing on uncertain objects help improve the model performances. 

\subsection{Analyses}
\subsubsection{Defining Target Objects by Uncertainty.}
\begin{figure*}[t]
    \begin{minipage}{1\textwidth}
    \centering
    \includegraphics[width=1.0\linewidth]{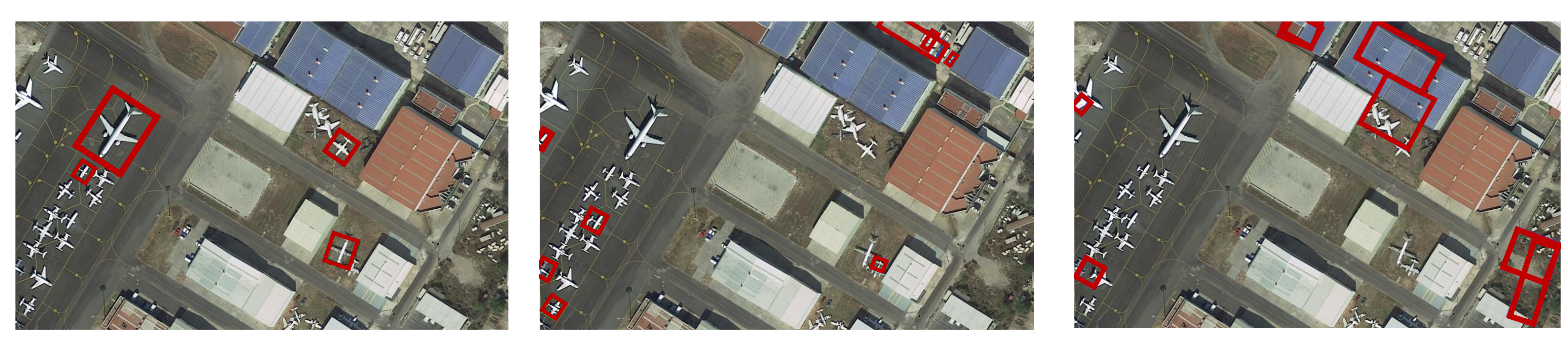}
    \end{minipage}
    \caption{\textbf{Visualization of Objects with Uncertainties.} We visualize detected objects with lowest, intermediate, and highest uncertainties in left, center, and right parts. The larger the uncertainties, more inaccurate detection results are.}
    \label{fig:uncertainty_visualization}
\end{figure*}
Fig. \ref{fig:uncertainty_visualization} illustrates initially detected objects along with their uncertainties. As shown in Fig. \ref{fig:uncertainty_visualization}, detected objects with the lowest uncertainties are accurate, whereas, objects with the highest uncertainties are completely wrong. Therefore, we only target objects with medium uncertainties, which are neither perfectly detected, nor totally wrong.
We can take two advantages when focusing on improving uncertain objects: computational efficiency and performance gain. To analyze this, we compare our UAGDet to SGRN~\cite{SGRN}, which builds a context graph among objects for contextualizing them. Specifically, in SGRN~\cite{SGRN}, 50 edges per node are generated for every detected object. However, if SGRN~\cite{SGRN} is applied to DOTA~\cite{DOTA} with 1,250 initial object proposals, resulting graph contains 50$\times$1250 edges for considering all objects' pair-wise interactions. Compared to such an approach, our UAGDet always generates 50$\times$50 edges as a maximum under its analytical form, and usually generates between 50 to 1,000 edges in most cases. This is because we define the direction of edges only from source to target nodes, which results in much sparser graphs having advantages in terms of time and memory efficiency. Furthermore, we observe that uncertainty-aware graphs with edges from certain nodes to uncertain nodes are more valuable than pair-wise edges between all nodes in terms of performance. In particular, we compare the performances of two models -- 73.1\% for all pair-wise interactions, whereas 75.1\% for our uncertainty-based interactions, on which we observe that ours largely outperforms the baseline. Also, the model with pair-wise edges among all nodes (73.1\%) underperforms the RoITrans~\cite{RoITransformer} baseline (73.4\%) that does not leverage the relational knowledge between objects. This result confirms that information propagation from uncertain objects to certain objects can harm the original detection results, which we prevent by constructing the bipartite directed graph. 

\subsubsection{Spatial and Semantic Edges in Graphs.} 
\begin{figure*}[t]
    \begin{minipage}{1\textwidth}
    \centering
    \includegraphics[width=1.0\linewidth]{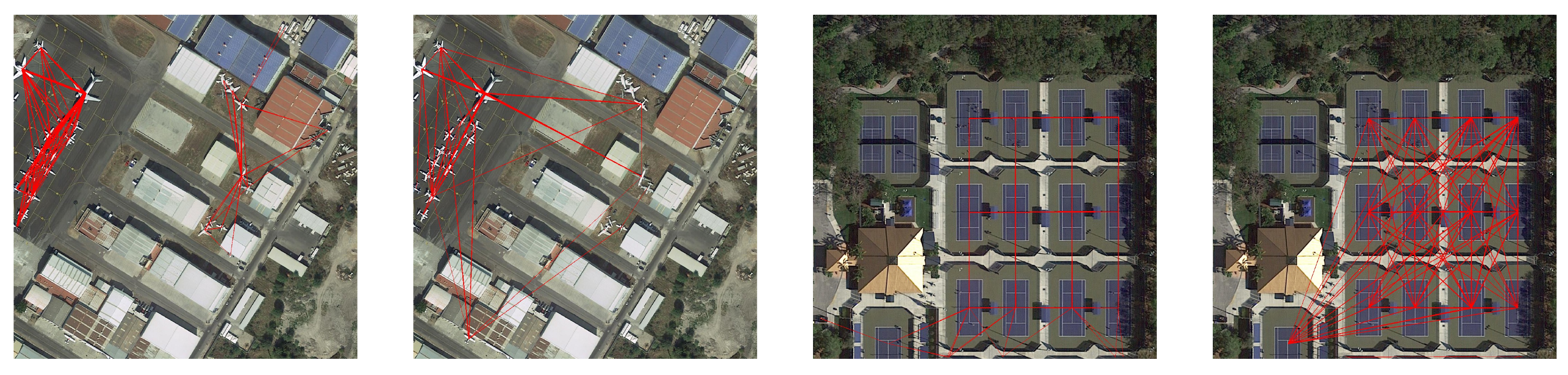}
    \end{minipage}
    \caption{\textbf{Visualization of Generated Object Graphs.} First and third images show edges generated by a spatial distance, while second and last images show edges generated by a semantic distance.}
    \label{fig:spatial_semantic_graph}
\end{figure*}

We use both the spatial and semantic similarities to decide if two nodes are related enough to be connected in a graph structure.
As depicted in Fig. \ref{fig:spatial_semantic_graph}, we observe that spatial and semantic distances play different roles when building a context graph. In particular, spatial distance measure is used to capture relationships within nearby objects. However, it is suboptimal to only consider geometric distances between objects for contextualizing various objects in the image (e.g., two planes located far away are not connected to each other in Fig.~\ref{fig:spatial_semantic_graph} only with spatial distances). Thus, 
\begin{wraptable}{r}{0.25\linewidth}
\centering
\caption{\small Results of spatial-only and semantic-only graph.}
\label{tab:spatial_semantic_ablation}
\begin{adjustbox}{width=1.\linewidth}
\begin{tabular}{@{}l c@{}}
\toprule
\small
Models & mAP \\
\midrule \midrule
\textbf{Ours} (Both Distances) &  \textbf{75.1} \\
Spatial Distance Only &  74.6 \\
Semantic Distance Only &  73.9 \\
\bottomrule
\end{tabular}
\end{adjustbox}

\end{wraptable}
for this case, considering semantic distance helps a model to capture semantically meaningful relationships even though objects are located far from each other, for example, tennis courts and planes in Fig.~\ref{fig:spatial_semantic_graph}. Experimentally, we observe that the model shows 74.6\% and 73.9\% performances in terms of mAP, when using spatial and semantic distances independently, which is shown in Table~\ref{tab:spatial_semantic_ablation}. Note that those two results are lower than the performance of 75.1\% which considers both spatial and semantic distances for edge generation. Therefore, such results empirically confirm that both spatial and semantic edges contribute to the performance gains, which are in a complementary relationship when contextualizing objects in the image. 

\section{Conclusion}
In this work, we proposed an Uncertainty-Aware Graph network for object DETection (UAGDet), a novel object detection framework focusing on relationships between objects by building a structured graph while considering uncertainties for representing and refining object features. In particular, we first pointed out the importance of object-level relationship which is largely overlooked in existing literature and then proposed to leverage such information by building an object graph and utilizing GNNs. Also, we considered uncertainty as a key factor to define the relationship between objects, where we transferred knowledge from certain objects to uncertain ones, refined only the uncertain object features, and regulated the loss value based on the uncertainty, for improving uncertain object predictions. Experimentally, our method obtained 2.32\% and 3.82\% performance improvements in the DOTA dataset.
\subsubsection{Acknowledgement}
This research was supported by the Defense Challengeable Future Technology Program of the Agency for Defense Development, Republic of Korea.
\newpage

%
%
\bibliographystyle{splncs04}
\bibliography{egbib}
\end{document}